%% file: sample-acmsmall.tex
\documentclass[format=acmsmall, review=false, screen=true]{acmart}

\usepackage{booktabs} 

\acmJournal{TWEB}
\acmVolume{9}
\acmNumber{4}
\acmArticle{39}
\acmYear{2018}
\acmMonth{6}
\copyrightyear{2018}

\setcopyright{acmlicensed}

\acmDOI{0000001.0000001}

\received{August 2017}
\received[revised]{May 2018}
\received[accepted]{September 2018}

\usepackage{graphicx}   
\usepackage[english]{babel}
\usepackage{lipsum} 
\usepackage{tikz}
\usepackage{multirow}
\usepackage[utf8]{inputenc}
\usepackage{hyperref} 
\usepackage{wrapfig}
\usepackage{enumitem}
\usepackage{algorithm}
\usepackage[noend]{algpseudocode}
\usepackage{amsmath}
\usepackage{chngcntr}
\usepackage{float}

\usetikzlibrary{
    shapes.geometric,
    positioning,
    fit,
    calc
}
\tikzset{
    dia/.style={
        shape=diamond,
        minimum size=2em,
    },
    dia cross/.style={
        dia,
        append after command={
            \pgfextra
                \draw[shorten >=\pgflinewidth,shorten <=\pgflinewidth]
                                   (\tikzlastnode.west) -- (\tikzlastnode.east);
                \draw[shorten >=\pgflinewidth,shorten <=\pgflinewidth]
                                 (\tikzlastnode.north) -- (\tikzlastnode.south);
            \endpgfextra
        }

    }
}

\makeatletter
\def\BState{\State\hskip-\ALG@thistlm}
\makeatother

\begin{document}
\title[Kurdish Transliteration]{A Rule-based Kurdish Text Transliteration System}  

\author{Sina Ahmadi}
\orcid{0000-0001-7904-6551}
\affiliation{%
  \institution{Paris Descartes University}
  \streetaddress{45 rue des Saints Pères}
  \city{Paris}
  \postcode{75006}
  \country{France}}

\begin{abstract}
In this article, we present a rule-based approach for transliterating two mostly used orthographies in Sorani Kurdish. Our work consists of detecting a character in a word by removing the possible ambiguities and mapping it into the target orthography. We describe different challenges in Kurdish text mining and propose novel ideas concerning the transliteration task for Sorani Kurdish. Our transliteration system, named \textit{Wergor}, achieves 82.79\% overall precision and more than 99\% in detecting the double-usage characters. We also present a manually transliterated corpus for Kurdish.
\end{abstract}

%
%


%
\setcopyright{acmcopyright}
\acmJournal{TALLIP}
\acmYear{2018} \acmVolume{1} \acmNumber{1} \acmArticle{1} \acmMonth{12} 



\maketitle

\renewcommand{\shortauthors}{Sina Ahmadi}

\input{samplebody-journals}

\end{document}

%% file: samplebody-journals.tex
\section*{Introduction}

Kurdish is an Indo-European language with a majority of speakers in the Kurdish regions of Iran, Iraq, Turkey and Syria. Although it is spoken by 20 to 30 million people \cite{Kreyenbroek2005,hassani2016automatic}, Kurdish language is considered as a less-resourced language. In 2016, Google added 13 new languages to its online automated translation tool, Google Translate, among them Kurdish (for the time being, only Kurmanji dialect). One of the main reasons of this delay, in comparison to some other languages with less users for whom the same service was provided earlier, is the lack of parallel corpora, online resources and language processing tools \cite{BESACIER201485}. 

Regarding the area and the extent to which Kurdish orthographies are applied, one should confess that still integrity in writing Kurdish has not been achieved. The difference of orthographies naturally results in the distinction of produced textual sources and adds to the gap between the dialects and thus scatters readers. Despite the fact that Kurdish Academy of Language introduced Unified Kurdish Alphabet \textit{Yekgirt\'{u}} in response to this problem \cite{WinNT3orth}, no standard orthography is popularly accepted considering all the challenges and the diversity of the dialects. Aware of this problem, Kurdish intellectuals have emphasized on the unification of the orthographies \cite{hassanpour1992nationalism}.

In this article, we are focusing on the challenges of transliteration of the two mostly used orthographies, Arabic-based and Latin-based, for Sorani Kurdish. Transliteration is a mapping from one system of writing into another, typically grapheme to grapheme \cite{Knight:1998:MT:972764.972767}. Given $w_{input}={c_1, c_2,...,c_n}$ in the orthography $A$, a transliteration task consists of mapping each character of the word to an equivalent character in the orthography $B$ which yields $w_{output}={c_1, c_2,...,c_m}$. This juxtaposition is not always straightforward. In the case of Sorani Kurdish, the Latin-based and the Arabic-based orthographies are not completely identical in terms of characters representation. Although confronting the problem of normalization in Kurdish seems to be addressed already in some of the previous researches such as \cite{esmaili2012challenges}, \cite{esmaili2014towards} and \cite{aliabadi2014towards} as a partial task, a solution has not been proposed for transliteration task so far. For instance, in a recent work by Hassani \cite{hassani2017kurdish}, transliteration has been mentioned implicitly as one of the tasks, but no detail has been reported concretely. 

The task of transliteration is one of the fundamental elements in many NLP applications such as statistical machine translation, terminology extraction, cross-lingual data linking and so forth. Transliteration can be done with phoneme-based or  grapheme-based models for which the latter has been shown to perform better than the first one \cite{al2002machine}. Kashani et al. \cite{kashani2007automatic} and Al-Onaizan and Knight \cite{al2002machine} use grapheme-based model, and Stalls and Knight \cite{stalls1998translating} and Pervouchine et al. \cite{pervouchine2009transliteration} use the phoneme-based approach. Since there are a few languages with manually labelled transliteration pairs (a word and its transliteration), some studies such as \cite{sajjad2017statistical,sajjad2011comparing,noeman2010language} have been focused on transliteration mining which consists of automatically extracting transliteration pairs from a noisy list of transliteration candidates.

The rest of the paper is organized as follows: First, we provide a description about Kurdish writing systems in section \ref{sec0}. In section \ref{sec1} we focus on the challenges of Sorani Kurdish transliteration in the Arabic-based (also referred to as "Persian-Arabic") and Latin-based orthographies. In section \ref{sec2} we present the rule-based techniques used in Wergor\footnote{"Wergor", pronounced as "wargor", is composed of "wer"-- a Kurdish prefix related to \textit{transformation}, and "gor"-- the stem of "goran" meaning\textit{ to change}. We coined this word for "transliterater" similar to the Kurdish word "wergêr" meaning \textit{translator}.}. This section includes our rule-based methods to solve the present challenges. Section \ref{sec_experiments} is devoted to the tests and experiments on the algorithms. In this section we describe our manually transliterated data set. Finally in section \ref{sec_final} our work is concluded and some ideas are proposed for future works. 

\section{Kurdish Writing Systems}
\label{sec0}
Nowadays Kurdish is written in quite several orthographies adopted from other languages and thus applied to it\cite{WinNT1orth}. Although debate on what orthography to apply yet remains, Latin-based orthography (henceforth referred to as \textit{LbO}) and Arabic-based orthography (henceforth referred to as \textit{AbO}) are among the most popular ones which are respectively mostly used for the \textit{Kurmanji} dialect and the \textit{Sorani} dialect of Kurdish. In addition to these two main dialects, \textit{Hawrami} and \textit{Kalhor} are also written in the AbO. These orthographies are based on the phonetics of the language \cite{WinNT2hist}.

In order to provide a common description about Kurdish orthographies, and avoid inconsistent descriptions, mainly in \cite{wahbi, hejar, w.m.thackstonSo2006, thackston2}, we have used the description in \cite{celadet2} for the LbO and the presented characters in \cite{blau} for the AbO. Although some of the characters may have other usages in other descriptions, these two references are mostly well-known for Kurdish writers. Table \ref{fig:tableau_des_lettres} shows the characters in these orthographies in comparison to one another. In the case a character does not exist for a given phoneme, the case is coloured in grey. We encourage future researchers to use the selected Latin-based orthography as it does not have any ambiguity.  

In the early stages of development of text processing tools for Kurdish, some fonts have been introduced to Kurdish users. \textit{Dilan fonts}, \textit{Ali fonts}, \textit{Zanest fonts} and \textit{Rebaz fonts} were among the most well-known fonts. These fonts were mainly based on the Persian and the Arabic keyboards and did not support Unicode. Fortunately, the existing characters in the Kurdish orthographies are completely supported by the Unicode standard. In the most recent development, the \textit{Kurditgroup} keyboard is proposed based on the Unicode characters which is widely used by most of Kurdish users\footnote{ \url{https://kurditgroup.org/downloads}}. We have also used this keyboard in our study.

\begin{table}[t]\centering
\includegraphics[scale = 0.51]{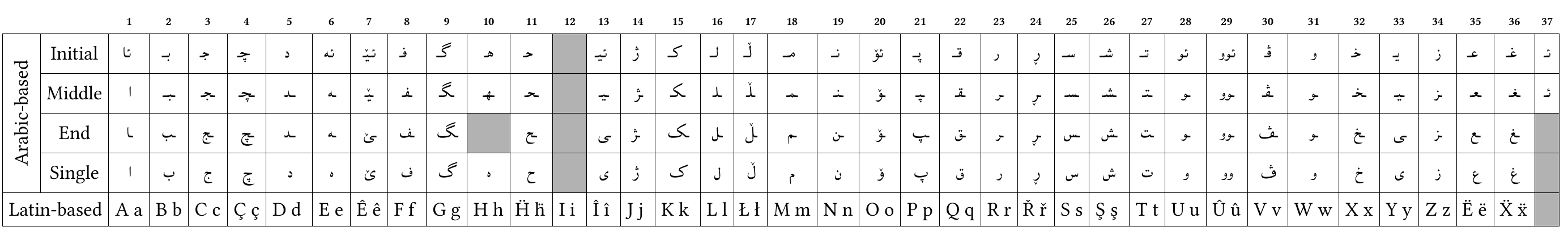}
\caption{Comparison of the Latin-based and the Arabic-based orthographies}
\label{fig:tableau_des_lettres}
\end{table}

\section{Kurdish Text Normalization Challenges}
\label{sec1}

For the current Arabic-based and Latin-based orthographies, we can classify the normalization challenges in 3 categories:

\subsection{Characters used to represent more than one phoneme} This is the case of "\includegraphics[scale = 0.02]{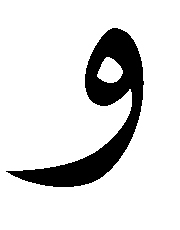}" and "\includegraphics[scale = 0.02]{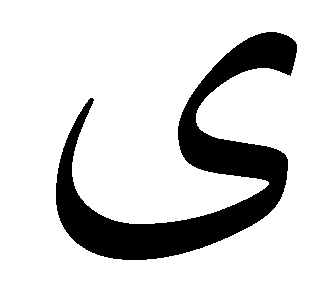}" in the AbO which may be transliterated respectively as \{"w" or "u"\} and \{"y" or "\^{i}"\} in the LbO. For instance, the word "\includegraphics[scale = 0.02]{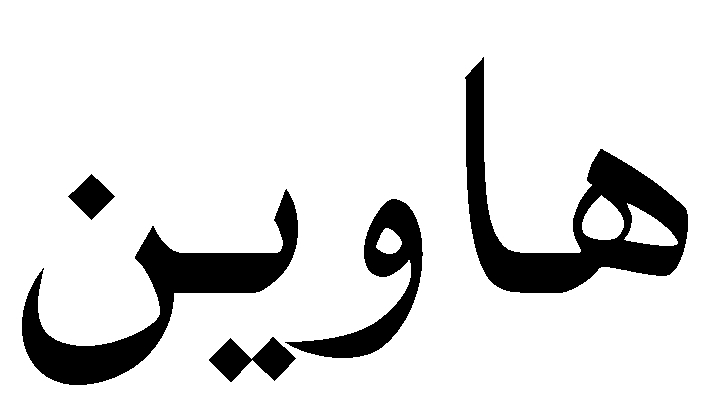}" could have 4 possible transliterations considering different mappings "\includegraphics[scale = 0.02]{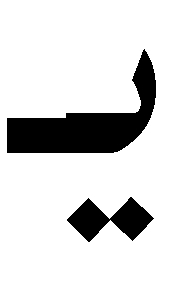}" $\rightarrow$  \{"y", "î"\} and "\includegraphics[scale = 0.02]{images/jpg/w.jpg}" $\rightarrow$  \{"w", "u"\}: "hauîn", "hauyn", "hawîn", "hawyn", for which "hawîn" is the correct form. Despite the visual similarity of "\includegraphics[scale = 0.03]{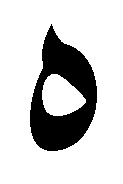}" as the equivalent of "h" and "e" in LbO, this character is not in the same category with "\includegraphics[scale = 0.02]{images/jpg/w.jpg}" and "\includegraphics[scale = 0.02]{images/jpg/y.jpg}" having different codes in Unicode.

\subsection{Characters with no equivalent in the other orthography} This is the case of "\includegraphics[scale = 0.02, trim =4mm 0mm 0mm 0mm]{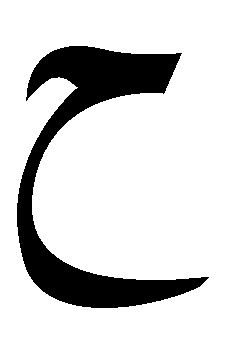}", "\includegraphics[scale = 0.02, trim =0mm 2mm 0mm 0mm]{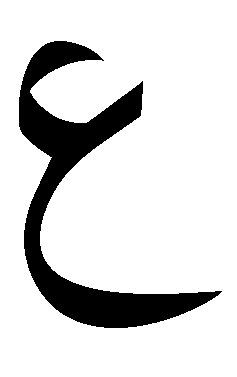}", "\includegraphics[scale = 0.02, trim =0mm 2mm 0mm 0mm]{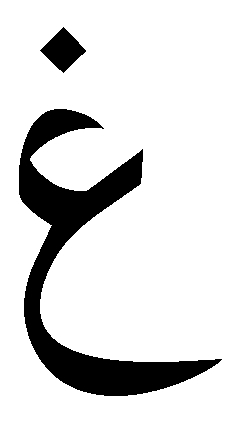}", "\includegraphics[scale = 0.02, trim =0mm 2mm 0mm 0mm]{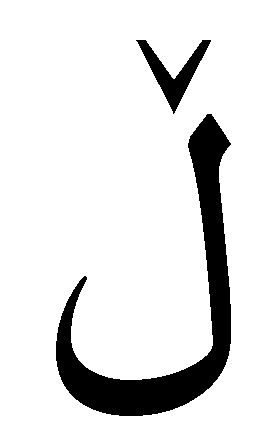}" and "\includegraphics[scale = 0.02, trim =0mm 2mm 0mm 0mm]{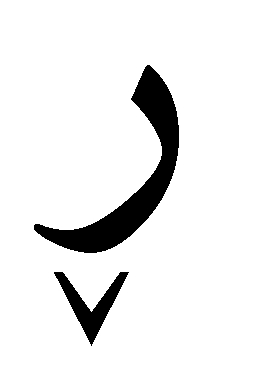}" characters in the AbO for which there is no equivalent in the LbO. A specific case, however, is the case of \textit{Bizroke}. Bizroke (which literally means "the little furtive") is represented by "i" in the LbO while it is totally ignored in the AbO. For example, the word "\includegraphics[scale = 0.025]{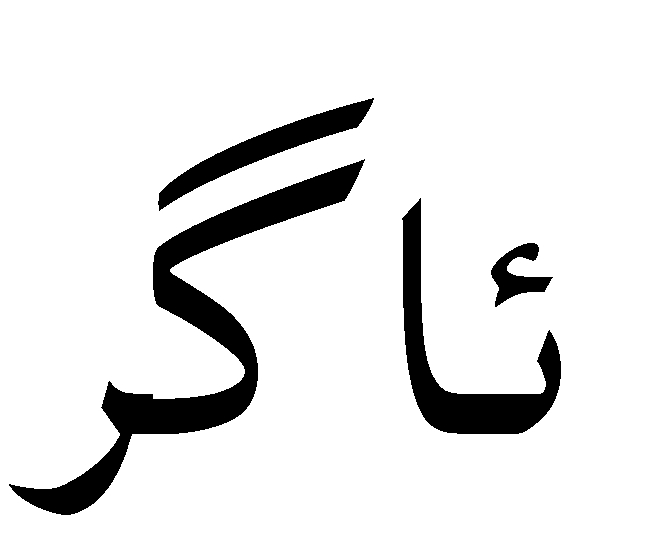}" may be transliterated as "agr" which is not correct since the Bizroke between "g" and "r" can not be represented in the AbO. The correct form is "agir". Having said that native speakers pronounce Bizroke while speaking, even if it does not exist in the Arabic-based orthography \cite{mccarus1958kurdish}.

\subsection{Unicode assignments of the Arabic-based Kurdish alphabet}
\label{unicodeassi}

The potential sources of ambiguity in the assignment of the characters of the current Kurditgroup keyboard is as follow:
 
\begin{itemize}

\item Some of the Arabic characters have similarities in form, but they have different Unicodes, e.g. "\includegraphics[scale = 0.7, trim =0mm 2mm 0mm 0mm]{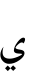}" (\texttt{U+064A}) instead of "\includegraphics[scale = 0.7, trim =0mm 1mm 0mm 0mm]{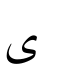}" (\texttt{U+06CC}) for \{"î", "y"\} and "\includegraphics[scale = 0.7, trim =0mm 1mm 0mm 0mm]{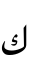}" ( \texttt{U+0643}) instead of "\includegraphics[scale = 0.7, trim =0mm 2mm 0mm 0mm]{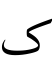}" (\texttt{U+06A9}) for "k" in the LbO.

\item Although "\includegraphics[scale = 0.03]{images/jpg/e.jpg}"(\texttt{U+0647}) as "h" is a connecting character when placed at the end of a word, it seems visually identical to "\includegraphics[scale = 0.03]{images/jpg/e.jpg}"(\texttt{U+06d5}) that represents "e". For instance, the final "\includegraphics[scale = 0.03]{images/jpg/e.jpg}" in "\includegraphics[scale = 0.6, trim =0mm 1mm 0mm 0mm]{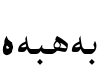}" ("behbeh") is not connected to the previous character which shows that the final "\includegraphics[scale = 0.03]{images/jpg/e.jpg}" is "h". This is not a source of ambiguity in terms of normalization since the two possible forms of "\includegraphics[scale = 0.03]{images/jpg/e.jpg}" have different Unicodes. Some suggest that "\includegraphics[scale = 0.03]{images/jpg/e.jpg}" as "h" be marked using a zero-width non-joiner character (\texttt{U+200C}) or an \textit{en dash} (\texttt{U+2013}). Such words ending with "h" phoneme are quite rare in Sorani Kurdish.

\item Although "û" in the LbO is a single character with a unique Unicode (\texttt{U+00FB}), the equivalent character "\includegraphics[scale = 0.02, trim =0mm 2mm 0mm 0mm]{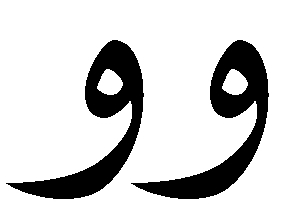}" in the AbO is created by a double "\includegraphics[scale = 0.02, trim =0mm 2mm 0mm 0mm]{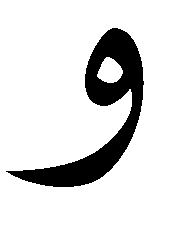}". The usage of two characters to represent another character is far problematic than a simple replacement since some of the words are preceded or succeeded by a similar character. For instance, the double "\includegraphics[scale = 0.02, trim =0mm 2mm 0mm 0mm]{images/jpg/vowels/w.jpg}" in  words like "\includegraphics[scale = 0.6, trim =0mm 2mm 0mm 0mm]{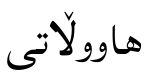}" and "\includegraphics[scale = 0.6]{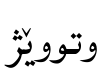}" may be transliterated respectively as "haûłatî" instead of its correct form "hawwiłatî" and "witûej" instead of its correct form "witûwêj". In a similar way, some have proposed using "ll" and "rr" to represent "\includegraphics[scale = 0.7, trim =0mm 2mm 0mm 0mm]{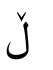}" and "\includegraphics[scale = 0.7, trim =0mm 2mm 0mm 0mm]{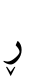}" in the LbO \cite{blau1999manuel}. Consequently, it would be the same case for such usages.   

\end{itemize}

\begin{table}[h]
\centering
\scalebox{0.8}{
\begin{tabular}{||c | c |  c |  c||}
\hline
Word & Possible transliterations & Correct form & Challenge category \\
\hline \hline

\multirow{4}{*}{\includegraphics[scale = 0.7]{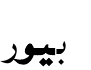}} & bîwr                     & \multirow{4}{*}{bîwir} &       "\includegraphics[scale = 0.02]{images/jpg/w.jpg}" $\rightarrow$  \{"w", "u"\}   \\
                            & bywr                    &                        &            "\includegraphics[scale = 0.02]{images/jpg/vowels/i_middle.jpg}" $\rightarrow$  \{"y", "î"\}\\
                            & bîur                    &                        &             \textit{Bizroke}, i.e., "i", not recognizable\\ 
                            & byur                      &                        &           
\\ \hline


\includegraphics[scale = 0.7, trim =0mm 2mm 0mm 0mm]{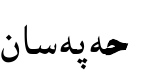} & hepesan                     & \"{h}epesan &  No character for "\includegraphics[scale = 0.7]{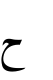}"        in the LbO
\\ \hline

\includegraphics[scale = 0.7, trim =0mm 2mm 0mm 0mm]{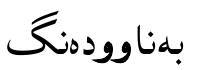} & benaûdeng                     & benawûdeng &  Double character for one character             
\\ \hline

\end{tabular}
}
\caption{Examples of different challenging categories in Sorani Kurdish text normalization. Challenging characters, if available, are bolded.}
\label{table_challenges_AbO}
\end{table}

Table \ref{table_challenges_AbO} shows some words in the AbO with the possible transliterated forms in LbO, the correct form for each word based on the reference orthography and the challenge category. Note that the possible transliterations are not essentially correct since they represent the possible mapping of the characters of one orthography to the other.

\section{Wergor system}
\label{sec2}

\begin{figure}[h]
	\centering
\begin{tikzpicture}
\node at (0, 1) {Text in the source orthography};
\end{tikzpicture}

\begin{tikzpicture}[
    every node/.append style={draw,minimum width=3em,minimum height=.5em},
    >=latex
]


\draw[->] (0, 1) -- (0, 0.3);

\node                (a) {Convert to Unicode UTF-8};

\node[below=.5 of a]    (c) {Detection of double usage characters};
\node[below=.5 of c] (d) {Detection of Bizroke};

\node[fit=(c)(d), label={[inner sep=0pt,minimum height=4ex]below:normalized text}] (cd) {};

\node[below=1 of cd] (e) {Character mapping};

\node[inner sep=1em,fit=(a)(cd)(e)] (left) {};

\foreach \pp/\pf/\pt in {--/a/c,
						--/c/d,
						--/d/e}
\draw[
    shorten >=\pgflinewidth,
    ->
] (\pf) \pp (\pt);
%



\draw[->] (0, -4) -- (0, -5);
\end{tikzpicture}

\begin{tikzpicture}
\node at (0, 9) {Text in the target orthography in Unicode};
\end{tikzpicture}

\caption{Wergor System architecture\label{system_arch}}
\end{figure}
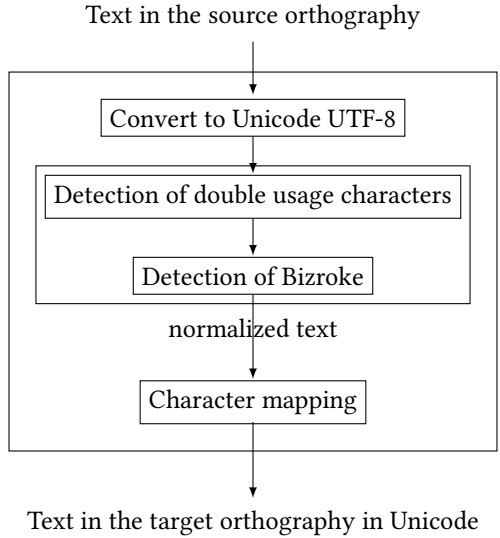

Figure \ref{system_arch} illustrates Wergor transliteration system architecture. The system normalizes a given text by preprocessing and unifying different forms of a character discussed in \ref{unicodeassi}. In this stage, Wergor yields the corresponding characters of the double-usage characters such as "\includegraphics[scale = 0.02]{images/jpg/w.jpg}" and "\includegraphics[scale = 0.02]{images/jpg/y.jpg}" and detects the possible presence of Bizroke in the AbO. Finally, the characters are mapped to the other orthography characters. According to this architecture, the system transliterates "\includegraphics[scale = 0.6, trim =0mm 2mm 0mm 0mm]{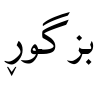}" from AbO into "bizguř" in the LbO by detecting the correct equivalent of "\includegraphics[scale = 0.02]{images/jpg/vowels/w.jpg}" as "u" and the correct position of Bizroke.

\newpage

%
%
%
%

Our method to solve the aforementioned challenges in Sorani Kurdish text processing follows the rules based on the phonological characteristics and the writing tradition. Some of the essential rules based on \citep{w.m.thackstonSo2006} that are applied in Wergor are as follow:

\begin{itemize}
\item If a word begins with a vowel, i.e., \{ "\includegraphics[scale = 0.02]{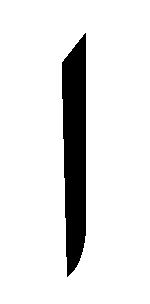}", "\includegraphics[scale = 0.02]{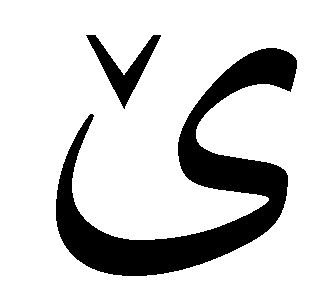}", "\includegraphics[scale = 0.02]{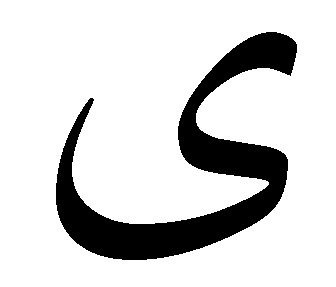}", "\includegraphics[scale = 0.02]{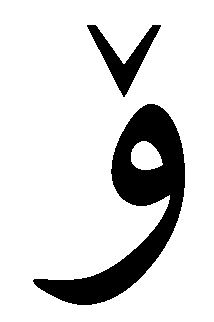}", "\includegraphics[scale = 0.02]{images/jpg/vowels/oo_char.jpg}", "\includegraphics[scale = 0.02]{images/jpg/vowels/w.jpg}", "\includegraphics[scale = 0.03]{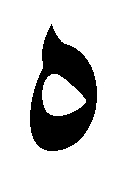}"\}, it is always preceded by "\includegraphics[scale = 0.7, trim =0mm 2mm 0mm 0mm]{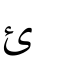}" in the AbO. This is the only usage of "\includegraphics[scale = 0.7, trim =0mm 2mm 0mm 0mm]{images/hemze.pdf}" (called \textit{Hamza}) as an auxiliary character and is only used in the AbO. 
\item Although "r" as the first phoneme in every word in the Sorani Kurdish is trilled, thus pronounced "ř", traditionally the non-trilled form "r" is used \cite{w.m.thackstonSo2006}. This rule is applied in the two orthographies. For instance, "\includegraphics[scale = 0.6, trim =0mm 1mm 2mm 0mm]{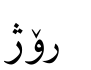}", "\includegraphics[scale = 0.6, trim =0mm 2mm 0mm 0mm]{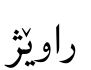}" and "\includegraphics[scale = 0.6, trim =0mm 2mm 0mm 0mm]{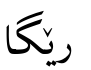}" are to be transliterated as "roj", "rawêj" and "rêga" respectively. 
\item No Sorani Kurdish word begins with "\includegraphics[scale = 0.02]{images/jpg/l.jpg} / ł" \cite{w.m.thackstonSo2006}.
\item Since in Sorani Kurdish a word has as many syllables as it has vowels, no two vowels can be in one syllable. Some of the frequent syllable structures in Sorani Kurdish are: V, VC, VCC, CV, CVC, CVCC, where V stands for vowel and C stands for consonant. In no syllable structure a vowel is preceded or succeeded by another vowel \cite{mccarus1958kurdish}. 
\end{itemize}

\begin{algorithm}[t]
\caption{Detection of "w/u" and "y/î" equivalents in the Arabic-based orthography}
\label{wydetector}

 \hspace*{-4.5cm} \textbf{Input}: Word \textit{W} containing the target char ("\includegraphics[scale = 0.02]{images/jpg/w.jpg}", "\includegraphics[scale = 0.02]{images/jpg/y.jpg}") \\
 \hspace*{-2cm} \textbf{Output}: Detected forms of "\includegraphics[scale = 0.02]{images/jpg/w.jpg}" as "w" or "u" and "\includegraphics[scale = 0.02]{images/jpg/y.jpg}" as "y" or "î" in \textit{W}.
 
\begin{algorithmic}[1]
\Procedure{TargetCharacterDetector}{W, TargetChar}

\State $\textit{length} \gets \text{length of W}$

\State $\textit{vowels} \gets$ ["i", "î", "u", "û", "\includegraphics[scale = 0.02]{images/jpg/vowels/e.jpg}",  "\includegraphics[scale = 0.02]{images/jpg/vowels/aa.jpg}", "\includegraphics[scale = 0.02]{images/jpg/vowels/o_char.jpg}", "\includegraphics[scale = 0.02]{images/jpg/vowels/ee.jpg}"]

\State $\textit{Hamza} \gets$ "\includegraphics[scale = 0.7, trim =0mm 2mm 2mm 0mm]{images/hemze.pdf}" 


\State $\textit{target\_char\_vowel} \gets$ the vowel form of \textit{TargetChar}
\State $\textit{target\_char\_consonant} \gets$ the consonant form of \textit{TargetChar}

\If{$W = TargetChar$}
\State \textbf{return} \textit{target\_char\_consonant}
\EndIf

\For{\textit{index} \texttt{$\gets$ 0 to }\textit{length}}
	\If{$W[index] = Hamza$ \& $W[index+1] = TargetChar$}
		\State $W[index+1] \gets$ \textit{target\_char\_vowel}
		\State $index \gets index+1$
	\Else
		\If{$W[index] = TargetChar$}
			\If{$index = 0$}
				\State $W[index] \gets$ \textit{target\_char\_consonant}
			\Else
				\If{$W[index-1] \text{ is in \textit{vowels}}$}
					\State $W[index] \gets$ \textit{target\_char\_consonant}
				\Else
					\If{$index+1 < length$}
						\If{$W[index+1] \text{ is in \textit{vowels}}$}
							\State $W[index] \gets$ \textit{target\_char\_consonant}
						\Else
							\State $W[index] \gets$ \textit{target\_char\_vowel}
						\EndIf
					\Else
						\State $W[index] \gets$ \textit{target\_char\_vowel}
					\EndIf
				\EndIf
			\EndIf
		\EndIf
	\EndIf
\EndFor	

\State Remove \textit{Hamza} in $W$
\State \textbf{return} $W$
\EndProcedure
\end{algorithmic}
\end{algorithm}

Using syllable structures pattern in Kurdish, we propose Algorithm \ref{wydetector} to detect double-usage characters "\includegraphics[scale = 0.02]{images/jpg/w.jpg}" and "\includegraphics[scale = 0.02]{images/jpg/y.jpg}". 
A character in its single form is considered consonant by default. The algorithm follows the same procedure for any of the target characters. 

Although the transliteration of Bizroke (i.e., "i") from the LbO to the AbO is by omitting it, it is challenging to find Bizroke in the inverse direction. Analyzing syllable structures, the only rule that we could rely on, is that in the CVC structure, if positioned as the first syllable, V is always Bizroke, e.g., "bira", "wirya", except the cases that the second consonant is "y" or "w", e.g., "kwêr", "dyar". Although it seems to be frequent to see Bizroke in the same pattern in the last syllables, e.g., "çirij", "kirdin", we could not use it as a rule. 

\section{Experiments}
\label{sec_experiments}

\subsection{Data set}

Among the 36 top ranked Kurdish websites, including news and media services, we have found only one site that uses AbO for both Sorani and Kurmanji dialects\footnote{Ranking based on Alexa \url{http://www.alexa.com}}. 18 websites use only LbO for Kurmanji and 29 websites use only AbO for Sorani. We found no Sorani website that uses LbO. 

In order to provide a resource for Kurdish transliteration, we propose Wergor corpus, to the best of our knowledge, as the first transliteration corpus for Kurdish. Our corpus consists of parallel transliterated texts from the two orthographies. This corpus can be used for other tasks in machine translation as well. 

\subsection{Results and Discussion}

Table \ref{ar2latin_results} shows the results of Wergor in transliterating our data set from the AbO to the LbO. Results of different tests are presented based on the correct and incorrect transliterations and the precision of the system is calculated as the the percentage of the correct transliterations. 

\begin{table}[h]
\centering
\scalebox{0.9}{
\begin{tabular}{ll|l|l|l|l|l|}
\cline{3-7}
\multicolumn{2}{l}{}                                                           & \multicolumn{2}{|l|}{Bizroke detection} & w/u detection              & y/î detection            & whole test set             \\ \hline
\multicolumn{1}{|l|}{\multirow{3}{*}{Prediction}} & Corrcet                     & \multicolumn{2}{l|}{721 / 1861}        & 2472 / 2480                & 4808 / 4850              & 5779 / 6980                \\ \cline{2-7} 
\multicolumn{1}{|l|}{}                            & \multirow{2}{*}{Incorrect} & last syllable                & other syllables            & \multirow{2}{*}{8 / 2480} & \multirow{2}{*}{48/4850} & \multirow{2}{*}{1201/6980} \\ \cline{3-4}
\multicolumn{1}{|l|}{}                            &                            & 286 / 1140         & 854 / 1140        &                            &                          &                            \\ \hline
\multicolumn{2}{|l|}{Precision}                                                & \multicolumn{2}{l|}{38.74\%}           & 99.67\%                    & 99.13\%                  & \textbf{82.79\%}                    \\ \hline
\end{tabular}
}
\caption{Arabic to Latin transliteration results}
\label{ar2latin_results}
\end{table}

In detecting the possible position of Bizroke, Wergor achieves 38.74\% precision and 100\% recall. Since the rule that we could apply in the current version of the system for detecting Bizroke only considers the first syllables, Wergor is not able to correctly find the position of Bizroke in the 1140 cases among 1861. In other words, the correct prediction refers to those words that have only one Bizroke and it is positioned in the first syllable. In the incorrect transliterations, in 286 cases Bizroke is in the last syllable and in 854 ones, it is in other syllables.

Evaluating the system on the double-usage characters, i.e., "\includegraphics[scale = 0.02]{images/jpg/w.jpg}" and "\includegraphics[scale = 0.02]{images/jpg/y.jpg}", shows a high precision of more than 99\% and a recall of 100\% since all relevant words were retrieved. Incorrectly transliterated words are mostly non-Kurdish words, e.g., "Claud" that are used in the original form in the manually transliterated data set, and proper nouns such as "Kurdistan" which are capitalized in the LbO. The AbO does not have capital letters. 

In the other hand, Wergor system achieves almost 100\% precision in transliterating the LbO into the AbO. Since the mapping of the LbO characters into the AbO ones is straightforward with no challenging characters, this precision is justifiable.

Figure \ref{example1} and \ref{example2} in Appendix \ref{appendix} shows two transliteration texts using Wergor.

\section{Conclusions and future work}
\label{sec_final}
In this paper, we propose a rule-based technique for Kurdish text transliteration. Kurdish confronts various challenges in transliterating its two popular orthographies, Arabic-based and Latin-based. In this article we described a method to solve these challenges using Wergor transliteration system. Although our system achieves 99\%  precision in transliterating double-usage characters ("\includegraphics[scale = 0.02]{images/jpg/w.jpg}", "\includegraphics[scale = 0.02]{images/jpg/y.jpg}"), it is less efficient in transliterating Bizroke, i.e., "i". In order to improve the current results, a bigger transliteration data set is required. We also believe that the phonological aspects of the language can be of help, which are not enough studied yet. Having the Wergor transliteration data set, we are currently interested in applying statistical methods for detecting Bizroke more efficiently. 

Our codes and corpus are available at \url{https://github.com/sinaahmadi/wergor}.

\bibliographystyle{unsrt}
{\small \bibliography{sample}}

\appendix
\section{APPENDIX}
\label{appendix}

\renewcommand\thefigure{\thesection.\arabic{figure}}    
\setcounter{figure}{0} 

\begin{figure}[h]
\centering
\includegraphics[scale = 0.7]{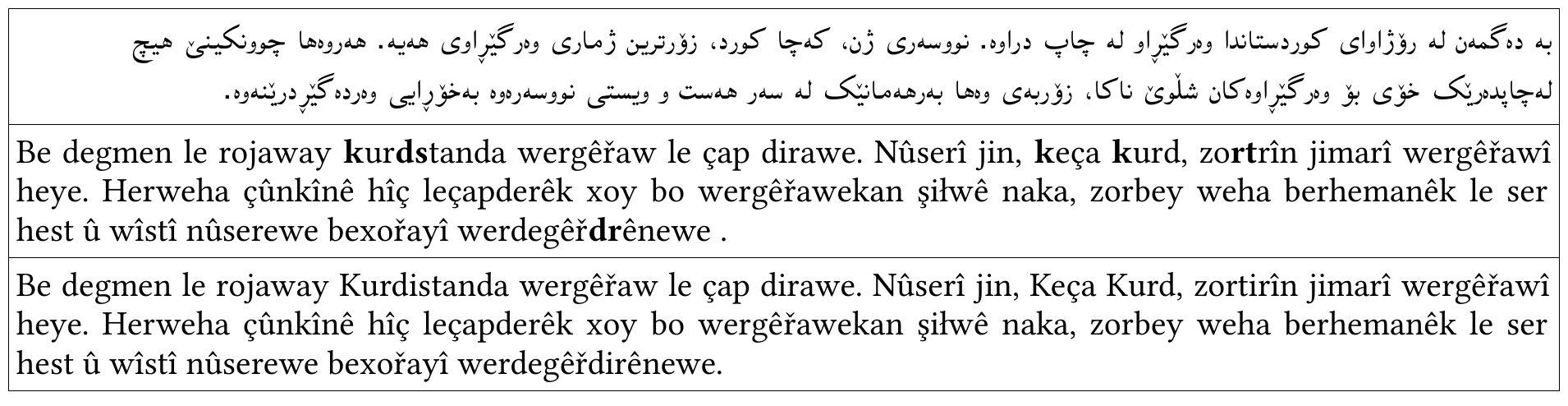}
\caption{Transliteration of an example text, in the first row, from the AbO to output text in the second row in the LbO. The manually transliterated text is shown in the last row. The errors are shown in bold. Both texts are in Sorani Kurdish language.}
\label{example1}
\end{figure}
\vspace{-1mm}
\begin{figure}[h]
\centering
\includegraphics[scale = 0.7]{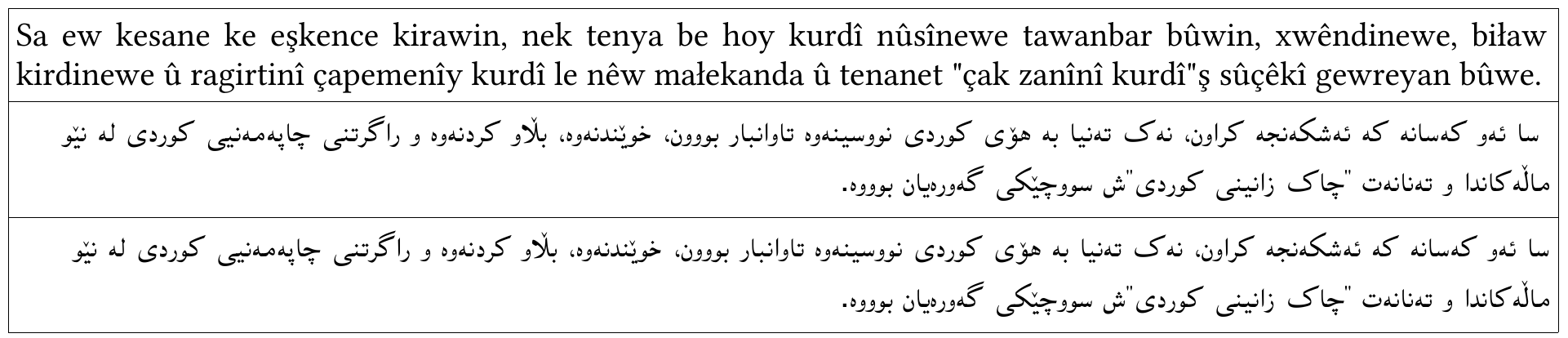}
\caption{Transliteration of an example text, in the first row, from the LbO to the output text in the second row in the AbO. The manually transliterated text is in the third row. No errors found.}
\label{example2}
\end{figure}